\documentclass[letterpaper, 10 pt, conference]{ieeeconf}
\IEEEoverridecommandlockouts
\usepackage{cite}
\usepackage{amsmath,amssymb,amsfonts}
\usepackage{graphicx}
\usepackage{textcomp}
\usepackage{xcolor}
\usepackage{amsmath}
\usepackage{graphicx}
\usepackage{subcaption}
\usepackage{algorithm}
\usepackage{algpseudocode}
\usepackage{siunitx,etoolbox}
\usepackage{subcaption}
\usepackage{hyperref}

\def\BibTeX{{\rm B\kern-.05em{\sc i\kern-.025em b}\kern-.08em
    T\kern-.1667em\lower.7ex\hbox{E}\kern-.125emX}}
\begin{document}

\title{Sensorless Estimation of Contact Using Deep-Learning \\for Human-Robot Interaction}

\author{Shilin Shan$^{1,*}$ and Quang-Cuong Pham$^{2}$ 
\vspace{-10pt}
\thanks{$^{1}$ Shilin Shan is with School of Mechanical and Aerospace
Engineering, Nanyang Technological University, Singapore (address: 50 Nanyang
Ave, Singapore 639798; phone: +65 6790 5568; e-mail: shilin.shan153@gmail.com)} 
\thanks{$^{2}$ Quang-Cuong Pham is with Eureka Robotics, and School of
Mechanical and Aerospace Engineering, Nanyang Technological University,
Singapore. (e-mail: cuong@ntu.edu.sg)} 
\thanks{$^{*}$ Corresponding author: Shilin Shan} }
\maketitle

\begin{abstract}
    Physical human-robot interaction has been an area of interest for decades.
    Collaborative tasks, such as joint compliance, demand high-quality joint
    torque sensing. While external torque sensors are reliable, they come with
    the drawbacks of being expensive and vulnerable to impacts. To address these
    issues, studies have been conducted to estimate external torques using only
    internal signals, such as joint states and current measurements. However,
    insufficient attention has been given to friction hysteresis approximation,
    which is crucial for tasks involving extensive dynamic to static state
    transitions. In this paper, we propose a deep-learning-based method that
    leverages a novel long-term memory scheme to achieve dynamics
    identification, accurately approximating the static hysteresis. We also
    introduce modifications to the well-known Residual Learning architecture,
    retaining high accuracy while reducing inference time. The robustness of the
    proposed method is illustrated through a joint compliance and task
    compliance experiment.
\end{abstract}

\vspace{-5pt}

\section{Introduction}

Physical human-robot Interaction has been an area of interest for many years. In
collaborative tasks such as joint compliance or task compliance, users are
allowed to change the robot configuration by simply pushing the robot link at
any position or dragging the robot end-effector. Consequently, external
force/torque sensing is crucial for achieving safe and dexterous collaboration.
Collaborative robots typically rely on costly and impact-prone joint torque
sensors to provide link-level external torque feedback. To address these
limitations, external torque approximation based on dynamics identification and
internal signals has been studied for years to reduce manufacturing costs and
enhance robot control schemes.

Numerous approaches have been proposed, falling into categories such as
model-based methods \cite{de2006collision, magrini2015control,
wahrburg2017motor, zhang2019sensorless, roveda2022robot}, data-driven model-free
methods \cite{yilmaz2020neural, sharkawy2020human}, and hybrid methods
\cite{liu2018end, hu2017contact, liu2021sensorless1}. While these methods have
demonstrated promising outcomes in various tasks like trajectory tracking,
wrench estimation, and assembly, their primary focus has been on robot dynamics
during continuous motions. However, a notable gap exists in the literature
concerning the hysteresis error that occurs at zero joint velocity. This
scenario is particularly relevant in tasks like joint compliance and task
compliance, which involve extensive joint status transitions from static to
dynamic and vice versa.

\begin{figure}[t]
    \centering
    \begin{subfigure}[b]{0.30\textwidth}
        \includegraphics[width=\linewidth]{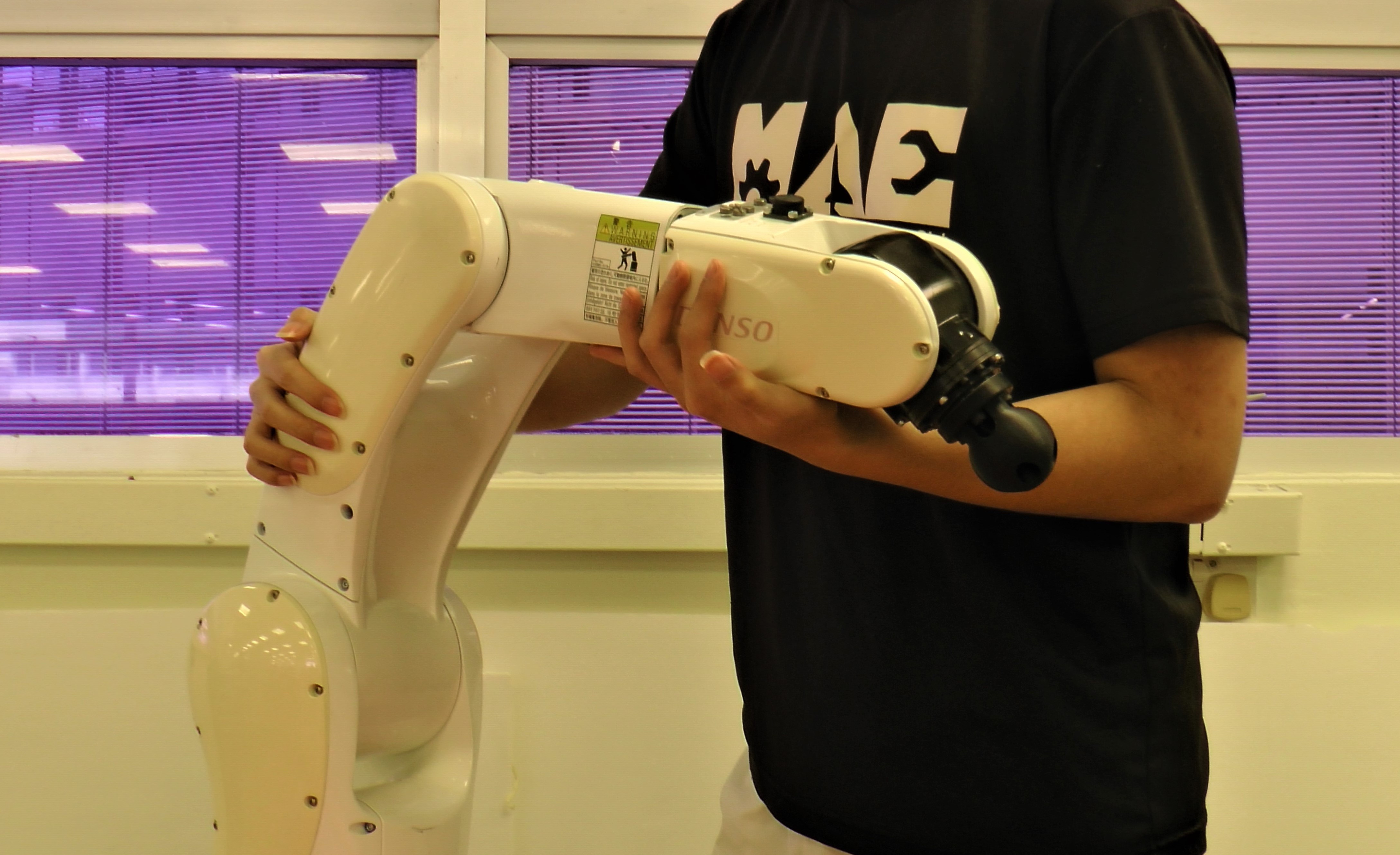}
        \label{fig:joint_compliance_snap}
    \end{subfigure}
    \begin{subfigure}[b]{0.17\textwidth}
        \includegraphics[width=\linewidth]{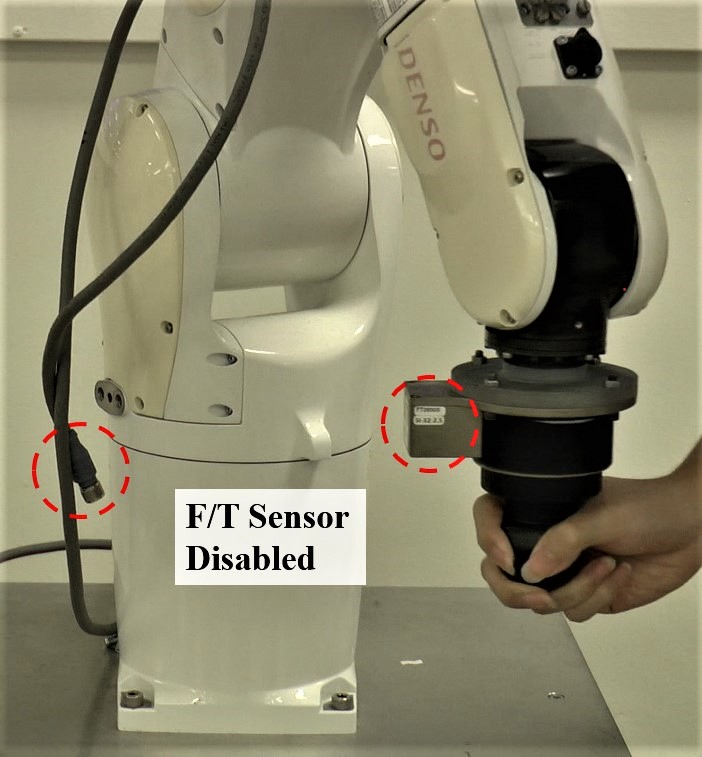}
        \label{fig:task_compliance_snap}
    \end{subfigure}
    \vspace{-8pt}
    \caption{Snapshot of joint compliance (left) and task compliance (right). A
    video clip is available in the supplementary or at:
    \href{https://youtu.be/Yrjf5tU94e8}{https://youtu.be/Yrjf5tU94e8} for higher
    resolution.}
    \label{fig:snaps}
    \vspace{-12pt}
\end{figure}

To address this challenge, we propose a Neural-Network-based method with a
special input scheme and an enhanced Neural Network structure to achieve
dynamics identification at the current level. Specifically, our main
contributions can be summarized as follows:
\begin{itemize}
    \item We designed a special input scheme for the Neural Networks that
    retains long-term temporal information of essential variables during
    execution. This is crucial for approximating static hysteresis, which
    requires non-local memory of joint states.
    \item We proposed modifications to the well-known Residual Learning method
    to make it adaptive to robot systems with high system frequency. Such
    modifications are effective in reducing model inference time while
    maintaining high accuracy.
\end{itemize}
We demonstrate the accuracy and reliability of the proposed method through joint
compliance based on admittance control and task compliance enabled by wrench
estimation.

The paper is organized as follows: In Section II, we discuss the related works
on dynamics identification, hysteresis approximation, and Neural Networks. In
Section III, we present the concepts for input design and modifications to
Residual Learning models. In Section IV, we discuss the data collection
procedures and modeling training results with ablation studies. In Section V, we
show the model performance in joint compliance and task compliance tasks.

\section{Related Works}

The study of dynamics identification has been conducted extensively for a
variety of applications, including collision detection, joint compliance, and
wrench estimation. Model-based approaches, predominantly augmented by
Generalized Momentum Observer (GMO) and the Kalman Filter, are designed to
approximate torque residuals using either measurable motor-driving torques
\cite{de2005sensorless, de2006collision, magrini2015control} or electrical
current coupled with predetermined motor constants \cite{gaz2018model,
wahrburg2017motor, wahrburg2015cartesian, roveda2017iterative}. However, relying
on current introduces additional challenges. Specifically, current signals tend
to be noisier than torque measurements, and the motor constants provided by
manufacturers are occasionally imprecise for certain robotic models
\cite{gautier2014global, xu2022robot}.

At low speeds or in static states, robots experience reduced measurement noise
but increased frictional issues. For instance, non-zero current readings can
occur at zero velocity, even without gravitational influence. Despite its
importance, only a handful of studies on dynamics identification have either
acknowledged or attempted to model this frictional behavior. In the work by
Wahrburg et al. \cite{wahrburg2017motor}, friction errors near zero velocity are
modeled as Gaussian noises and are mitigated through the use of Kalman Filter.
Meanwhile, Gaz et al. \cite{gaz2018model, gaz2019dynamic} approximate these
errors with a sigmoid function, accounting for the continuous transition from
negative to positive velocity. Comprehensive studies in
\cite{lampaert2002modification, al2005generalized, liu2021sensorless2} attribute
such errors to hysteresis and use the Maxwell Slip (MS) model for
identification, which demands long-term, non-local memory of joint states.
However, the static hysteresis in one joint appears to be affected by the
movements of other joints in our settings (detailed in Section V), posing
limitations for isolated hysteresis modeling methods.

Another approach to dynamics identification is model-free methods like
Neural-Network-based techniques, which have recently gained extensive attention.
Deep learning literature confirms that Neural Networks excel at modeling
nonlinear relationships when equipped with suitable architectures and unbiased
training data \cite{nielsen2015neural}. In the context of robot dynamics, neural
networks bypass the need for intricate mathematical analyses by learning optimal
solutions from unbiased datasets.

As a result, Neural-Network-based methods are being deployed to simplify or
improve the mathematical modeling of robot manipulators. Various NN
architectures, such as Multilayer Perceptron (MLP) \cite{sharkawy2020human,
kim2021transferable, liu2018end, hu2017contact, yilmaz2020neural}, Recurrent
Neural Networks (RNN) \cite{hanafusa2019external}, and Convolutional Neural
Networks (CNN) \cite{hanafusa2019external, lee2018interaction,
xia2021sensorless}, have been adapted for diverse tasks and settings. Notably,
Long Short-Term Memory (LSTM) networks have demonstrated effectiveness in
capturing and learning temporal patterns, making them a favored choice for
approximating hysteresis in various applications \cite{lim2021momentum,
wu2021hysteresis, hirose2017modeling}.

\begin{algorithm}[b]
    \caption{MD Update for Threshold $t_i$}\label{alg:MD}
    \begin{algorithmic}
        \For{$k = 1$ To Number-of-Joint}
            \If{$abs(\dot{q}[k]) \geq t_i[k]$}
            \vspace*{1pt}
                \State MD$[k] = \dot{q}[k]$
            \EndIf
        \EndFor
    \end{algorithmic}
\end{algorithm}

However, for applications involving joint and task compliance, the robot is
often required to remain stationary for extended periods without external
forces. Using LSTM in these settings poses challenges, especially at high system
frequencies (above 200Hz for industrial robots) and when approximating
hysteresis based on data from several minutes in the past. In this scenario, two
key issues arise: (i) Although LSTM is designed for long-term memory retention,
its capacity can wane when recalling information from thousands of frames back.
(ii) The recurrent nature of LSTM can lead to output drift, meaning that even
with a constant input, the estimation can fluctuate, undermining the model's
reliability.

To this end, we are motivated to design an input scheme that keeps a
non-vanishing memory of joint states while also ensuring stable estimations.
Drawing inspiration from the aforementioned MS model, we introduce in the next
section a combination of the proposed input scheme and modified Residual
Learning architecture based on MLP.

\section{Model Structures and Input Scheme}

\subsection{MLP and LSTM}

\begin{figure}[t]
    \vspace{5pt}
    \centering
    \begin{subfigure}[b]{0.24\textwidth}
        \includegraphics[width=\linewidth]{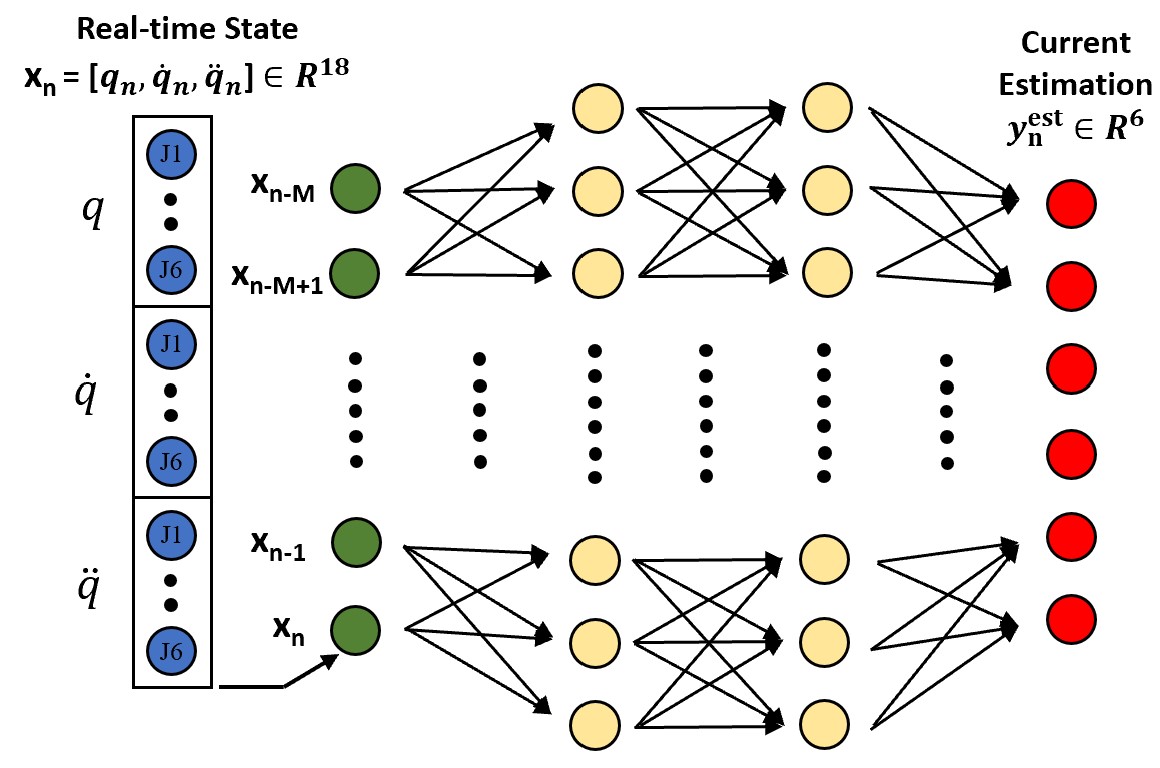}
        \caption{MLP}
        \label{fig:MLP_diagram}
    \end{subfigure}
    \begin{subfigure}[b]{0.22\textwidth}
        \includegraphics[width=\linewidth]{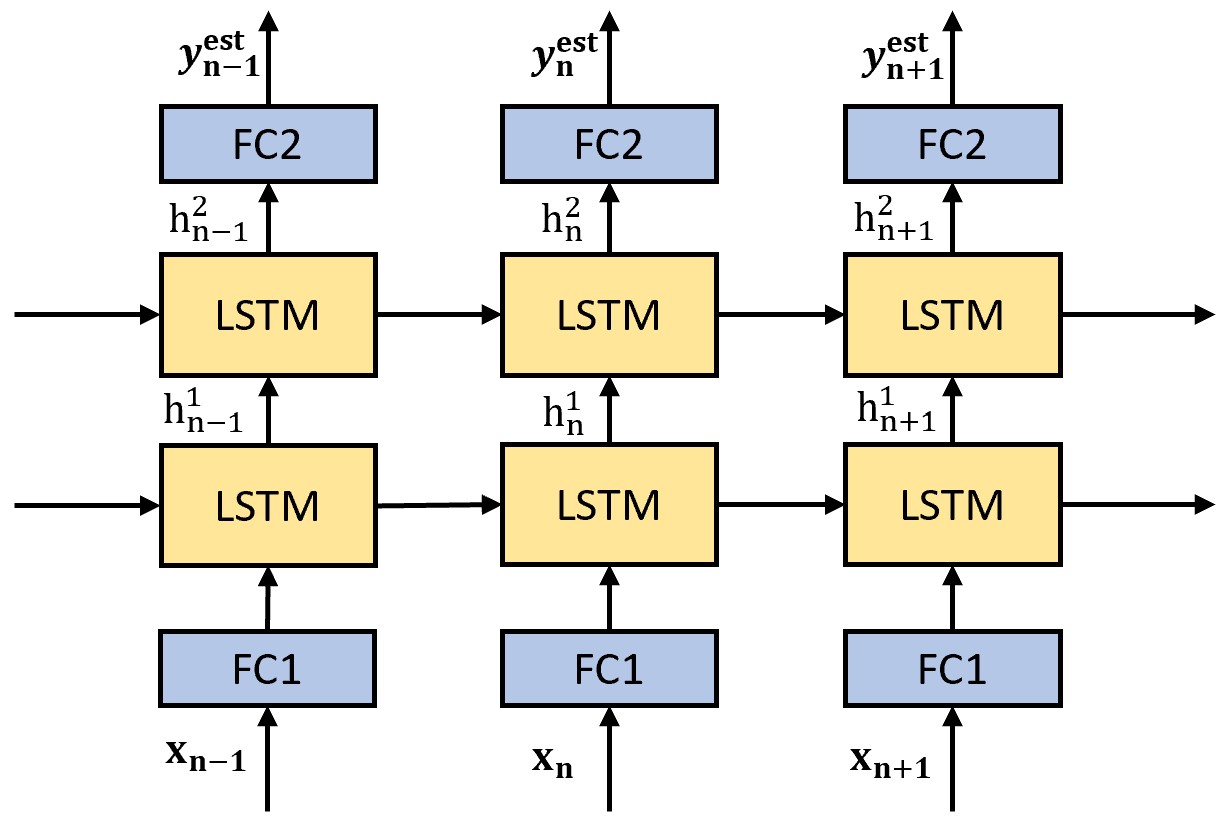}
        \caption{LSTM}
        \label{fig:LSTM_diagram}
    \end{subfigure}
    \caption{(a) The MLP example featuring two hidden layers, the notations $x$,
    $y$, $n$, and $M$ represent the input joint states, estimated currents,
    index of the instantaneous time step, and predefined short-term memory
    length, respectively. (b) The LSTM example with two LSTM cells, the symbol
    $h$ denotes the hidden states.}
    \label{fig:MLP_LSTM_diagrams}
    \vspace{-15pt}
\end{figure}

A standard MLP architecture is illustrated in Fig. \ref{fig:MLP_diagram},
consisting of an input layer, multiple fully-connected (FC) hidden layers, and
an output layer. While several studies \cite{yilmaz2020neural,
sharkawy2020human} have used instantaneous joint states as input, we enhance the
model's memory by including joint states from $M$ consecutive time steps
(short-term memory) in the input layer. This idea shares similarities with
approaches found in \cite{tran2020deep, zhang2021modeling}. Based on our
observations, this multi-frame input design results in improved estimation
accuracy and reduced noise.

The baseline LSTM model can be seen in Fig. \ref{fig:LSTM_diagram}. For each
recurrent unit, the input is propagated through an encoder, LSTM cells, then an
output decoder. Given the trivial patterns of joint states, we use FC layers for
encoding and decoding and one recurrent unit for each data frame. Further
details in the LSTM cells can be found in \cite{hochreiter1997long}. 

\subsection{Motion Discriminator Input Scheme}

The proposed input scheme introduces a global variable called the Motion
Discriminator (MD), designed to capture temporal information crucial for
enabling Neural Networks to differentiate between static and dynamic behaviors.
Specifically, for each frame of joint states, denoted by $x_n$, we define
multiple thresholds $T = (t_1, t_2, \ldots, t_I)$. Each threshold $t_i \in
\mathbb{R}^6$ (one element for each joint) is used to update the MD via a
thresholding function TH($\cdot$), as outlined in Algorithm \ref{alg:MD}.
Alternatively, the update law can be described as follows: if a joint's velocity
surpasses a given threshold, the joint is considered non-static and the MD is
updated with that velocity value. If the velocity falls below the threshold, no
update is made. As a result, when a joint's velocity nears zero, the MD retains
the signed value of the threshold. The complete MD is constructed by
concatenating $I$ vectors, each generated from a thresholding function.

\begin{figure}[t]
    \vspace{8pt}
    \centering
    \includegraphics[width=0.95\linewidth]{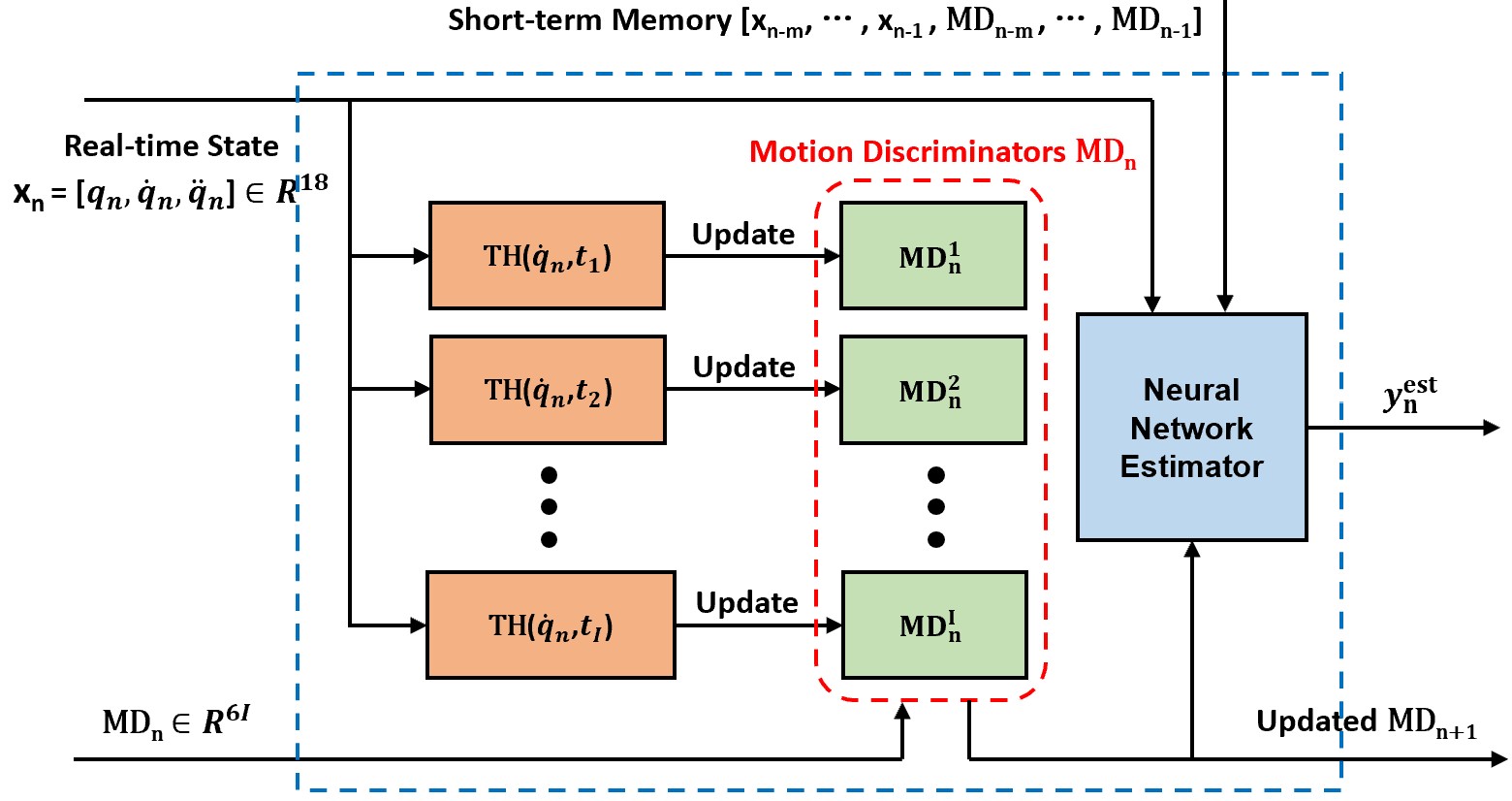}
    \caption{The block diagram for the Motion Discriminator input scheme.
    TH($\cdot$) and $t_i$ denote the thresholding functions and velocity
    thresholds.}
    \label{fig:MD_diagram}
    \vspace{-10pt}
\end{figure}

\begin{figure}[t]
    \vspace{8pt}
    \centering
    \includegraphics[width=0.45\linewidth]{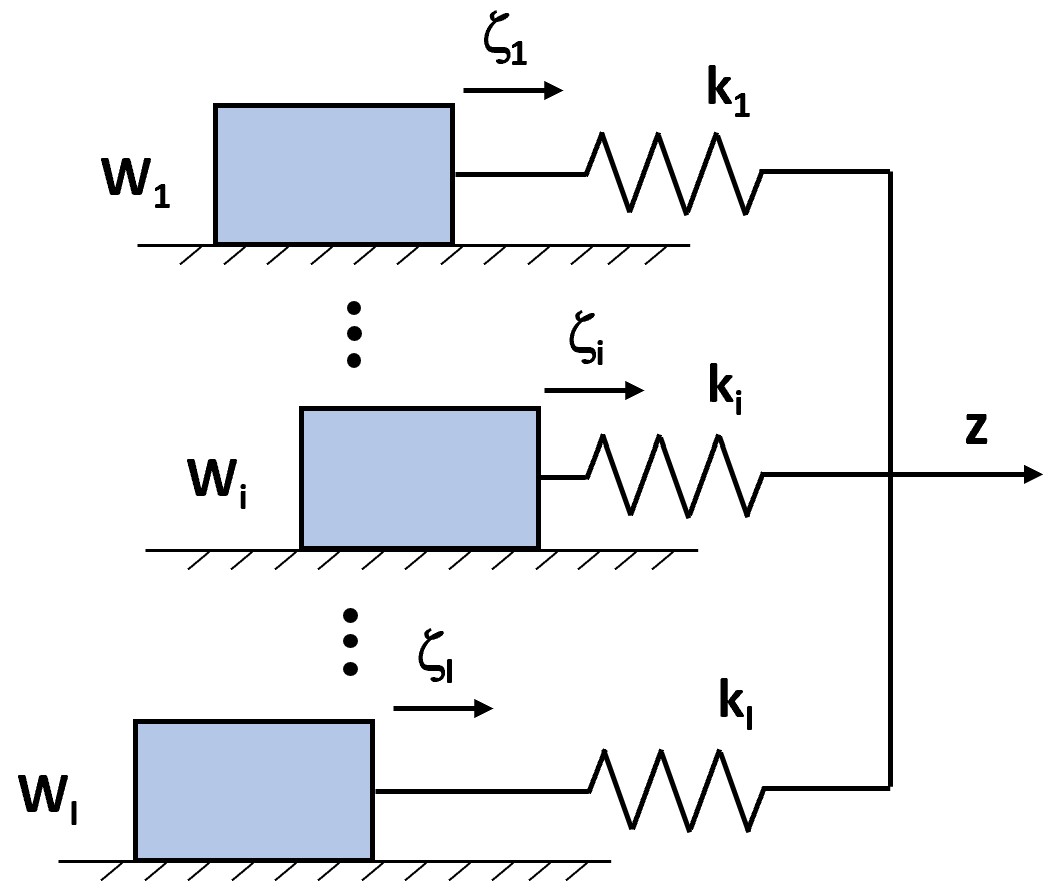}
    \caption{The Maxwell Slip model of $I$ elasto-slide elements.}
    \label{fig:MS_diagram}
    \vspace*{-5pt}
\end{figure}

As previously noted, the MD design draws inspiration from the Maxwell Slip (MS)
friction model. As visualized in Fig. \ref{fig:MS_diagram}, the macro friction
behavior is modeled by the parallel connection of $I$ elasto-slide elements. The
hysteresis friction of an individual element is formulated as follows:
\begin{equation}
    \vspace*{-2pt}
    \begin{aligned}
    \text{If } |z-\zeta_i|<\Delta_i \text{, then} \begin{cases}
        F_i = k_i(z-\zeta_i) \\
        \zeta_i = \text{constant}
    \end{cases}\\
    \text{else} \begin{cases}
        F_i = \text{sgn}(z-\zeta_i)W_i \\
        \zeta_i = z - \text{sgn}(z - \zeta_i)\Delta_i
    \end{cases}
    \end{aligned}
\end{equation}
where $z$ is the common displacement input, and $\Delta_i = {\scriptstyle
\frac{W_i}{k_i}}$, $\zeta_i$, $k_i$, and $W_i$ are the saturation displacement,
element position, stiffness, and saturation force of the element $i$. In this
formulation, $\Delta_i$ serves as a criterion to determine whether an element
should follow a sticking or a slipping model. The overall friction is given by
the sum of all elementary frictions, which can also be expressed in a
specialized manner to better clarify the input-output relationship:
\vspace*{-3pt}
\begin{equation}
    F_h = \sum_{i=1}^IF_i = \sum_{i=1}^IW_i\Phi_i(z_n,\zeta_{ni},\Delta_{ni})
    \vspace*{-3pt}
\end{equation}
where $\Phi(\cdot)$ is a nonlinear function described by (1). Given a constant
system frequency, utilizing the MD and joint states as NN inputs achieves two
key objectives: (i) Building multiple virtual elasto-slide elements, with their
sticking/slipping statuses discriminated by the thresholds $\Delta_i$ = $t_i/f$,
where $f$ is the system frequency; (ii) Providing the necessary input for MS
modeling, i.e., $z_n$ = $q_n$, $\Delta_{ni}$ = $sgn(\dot{q}_n)\Delta_i$, and
$\zeta_{ni}$ = $z_n-\Delta_{ni}$. 

\begin{figure}[t]
    \centering
    \begin{subfigure}[b]{0.40\textwidth}
        \includegraphics[width=\linewidth]{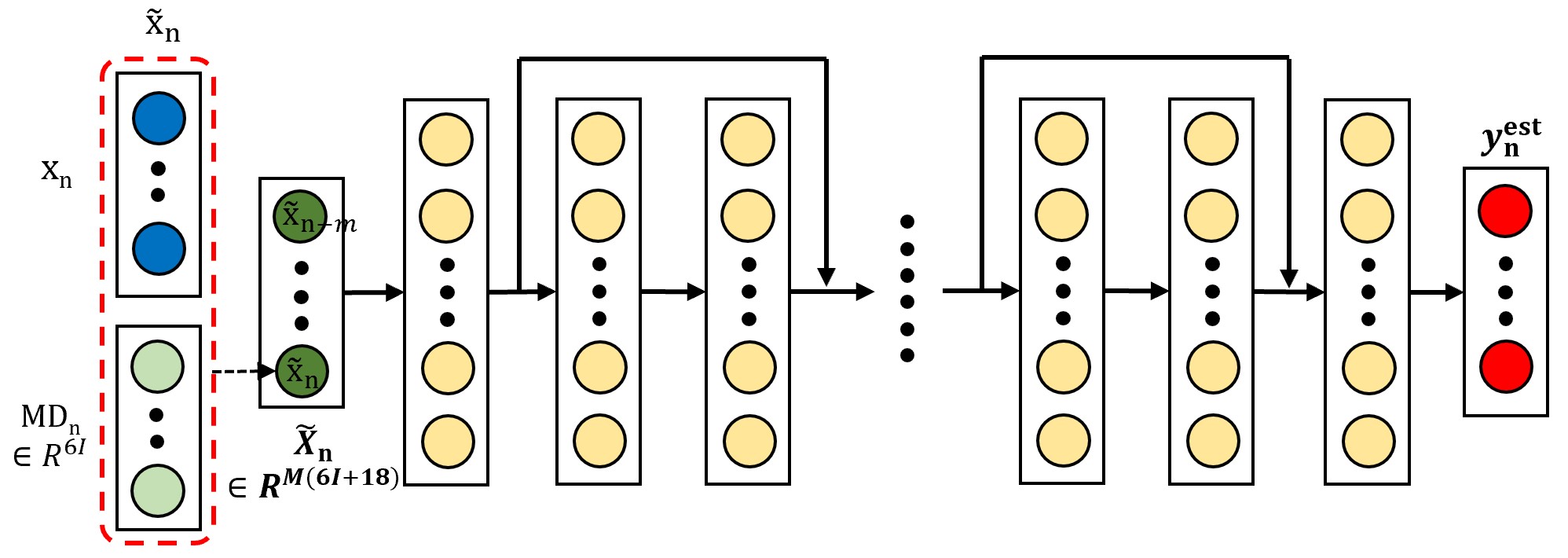}
        \vspace*{-20pt}
        \caption{}
        \label{fig:RL_diagram}
    \end{subfigure}
    \begin{subfigure}[b]{0.45\textwidth}
        \includegraphics[width=\linewidth]{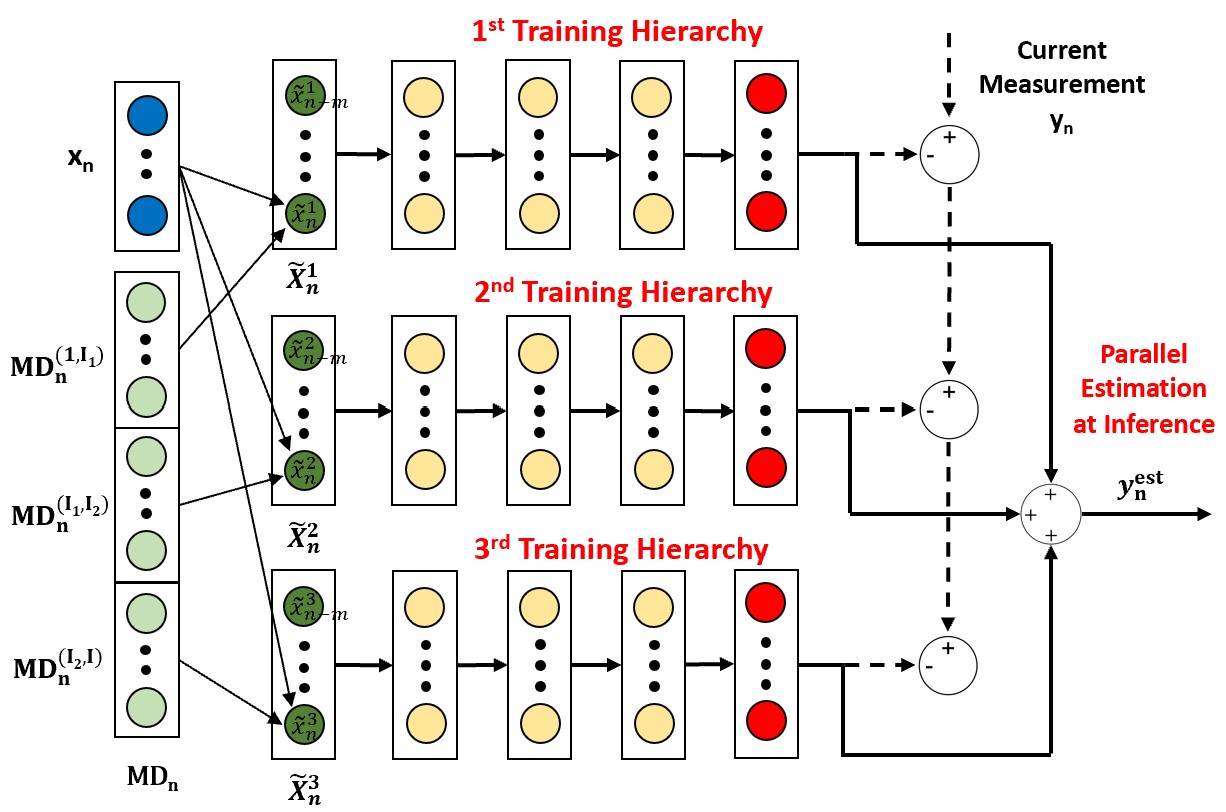}
        \vspace*{-20pt}
        \caption{}
        \vspace*{-3pt}
        \label{fig:HRL_diagram}
    \end{subfigure}
    \caption{(a) The classical Residual Learning (RDL) structure. $\tilde{x}$
    represents the combined input. (b) The Hierarchical Residual Learning (HRDL)
    architecture. The dashed and solid lines indicate the data flow in the
    training phase and during online inference, respectively. The superscripts
    of MD, e.g., ($I_1,I_2$), indicate the subset of MD with index from $I_1$ to
    $I_2$.}
    \label{fig:RL_HRL_diagrams}
    \vspace{-12pt}
\end{figure}

\subsection{Hierarchical Residual Learning}

With an appropriate input scheme, the Neural Network structure should also be
selected carefully to reduce inference time while ensuring accuracy. Residual
Learning \cite{he2016deep}, abbreviated as RDL in this paper, is known for
mitigating the vanishing gradient problem, thus enabling deep Neural Network
structures. When integrated with multi-frame and MD input schemes, an RDL model
can be designed for the dynamics identification purpose, as shown in Fig.
\ref{fig:RL_diagram}. However, for general-purpose applications lacking
additional computational resources like GPUs, a large number of input frames,
velocity thresholds, or hidden layers can result in prolonged inference times,
which is problematic for tasks requiring real-time force feedback.

We tackle this issue by transforming the basic Residual Learning architecture
into a Hierarchical Residual Learning (HRDL) design, illustrated in Fig.
\ref{fig:RL_HRL_diagrams}. During training, the first hierarchy processes joint
states, a subset of MD, and multi-frame short-term memory through a basic MLP
structure to yield an initial current estimation. A new dataset, comprising the
error between this estimation and ground truth, is then generated. The second
hierarchy, utilizing the same MLP structure but different subsets of MD, is
trained on this residual data to refine the initial estimation. Additional
hierarchies can be added in a similar fashion, each trained to estimate the
residual error of its predecessor. During online inference, all hierarchies
operate in parallel, taking advantage of the CPU's multi-core capabilities and
ensuring the inference time is bounded by the forward propagation time of a
single hierarchy.

This modification is not just applicable to dynamics identification problems; it
can also be effective for general tasks requiring both high accuracy and quick
inference from deep Neural Networks.

\section{Data Collection and Training}

\subsection{Training Data Collection}

In model-based methods, rigid body and friction dynamics equations are
predefined, allowing for minimal data collection through optimized exciting
trajectories \cite{gaz2019dynamic}. In contrast, data-driven model-free methods,
particularly those based on Neural Networks, make no prior assumptions, thus the
training results rely solely on the dataset distribution. To address this, we
collected data under two scenarios: first, where the robot follows continuous,
pre-planned trajectories for predefined tasks; and second, where it responds to
random external forces, leading to frequent shifts between static and dynamic
joint states.

\subsubsection{Continuous Trajectory Dataset}
The dataset's trajectories were generated using via-point path planning in the
task space. Joint velocity and acceleration were obtained from the first and
second derivatives of joint position. A third-order Butterworth Filter was
applied to minimally reduce signal noise. However, we retained some noise in the
dataset, as we believe the NN models should learn from noisy signals to enhance
its robustness. Joint states and currents were recorded at a frequency of 100Hz,
totaling 1,235,247 frames.

\subsubsection{Hysteresis-rich Trajectory Dataset}
Trajectories in this dataset were generated in the joint space to simulate the
on-off behaviors induced by external forces. The generation followed a
repetitive algorithm: (i) Random target positions were sampled for each of the
six joints; (ii) Joints moved toward these targets at random constant
velocities, pausing for 3 seconds after every 3 seconds of execution; (iii)
Joints stayed stationary upon reaching their goals until all goals were met.
This design aimed to enrich the dataset with interruptions, thereby offering
Neural Networks hysteresis information gathered from static scenarios. All data
were recorded at 100Hz, totaling 1,865,016 frames.

\subsection{Method Implementation}

For the proposed method, we implemented a 3-hierarchy HRDL-MD architecture, with
each hierarchy featuring 3 hidden layers of 256 neurons. The short-term memory
uses $M=5$. Through a series of screening experiments, we selected 15 velocity
thresholds ($I=15$) that yielded the highest accuracy. The detailed threshold
values and groupings are provided in Table \ref{table:HRL_threshold}.

\begin{table}[t]
    \vspace{7pt}
    \centering
    \setlength{\tabcolsep}{5.5pt}
    \begin{tabular}{c|c c c c c}
    \hline
    \multicolumn{6}{c}{HRDL Thresholds} \\ \hline
    Hierarchy & $t_1$ & $t_2$ & $t_3$ & $t_4$ & $t_5$ \\ \hline
    1 & 0.003 & 0.006 & 0.009 & 0.012 & 0.015 \\
    2 & 0.012 & 0.016 & 0.020 & 0.024 & 0.028 \\
    3 & 0.002 & 0.006 & 0.010 & 0.014 & 0.018 \\ \hline
    \end{tabular}
    \caption{Values and groupings of HRDL-MD thresholds.}
    \label{table:HRL_threshold}
    \vspace{-10pt}
\end{table}

For baseline comparisons, we used the model-based method proposed in
\cite{gaz2018model}. In particular, we identified dynamics parameters at the
current level using sinusoid exciting trajectories and applied the Generalized
Momentum Observer (GMO) for current residual approximation in the joint
compliance experiment in Section VI.

For the model-free methods, we implemented the basic MLP, LSTM and RDL models.
The detailed model depth and layer size are omitted due to limited space, but
can be inferred by comparing the inference times listed in Table \ref{table:2b}
to the HRDL-MD scheme. All models employed ReLU activation in fully connected
layers, except those preceding the output layer. LSTM layers utilized default
sigmoid and tanh activations. All models were trained using MSE loss and
optimized with the Adam algorithm. Notably, limiting LSTM input to 100-150
frames (1-1.5 seconds of memory) yielded optimal results, while extending input
beyond 200 frames reduced accuracy instead.

\section{Error Analysis with Ablation Study}

\begin{figure}[t]
    \vspace{5pt}
    \centering
    \includegraphics[width=0.95\linewidth]{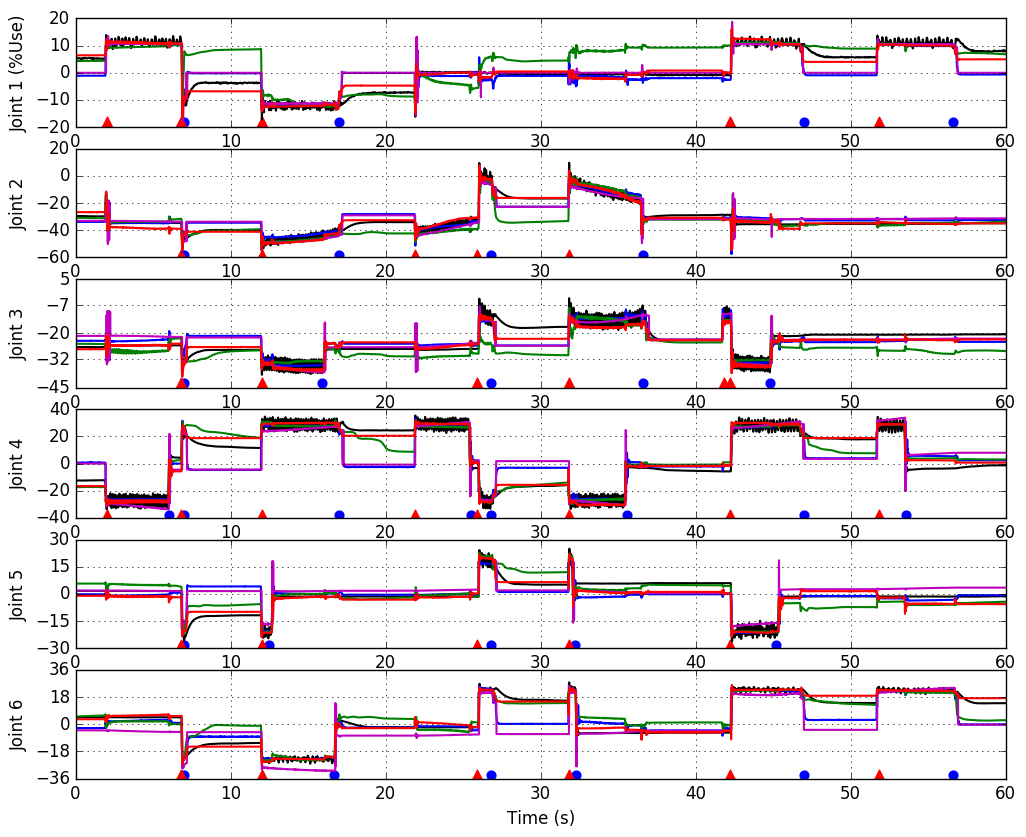}
    \vspace{-5pt}
    \caption{Comparison of proposed and baseline methods on the hysteresis-rich
     trajectories. The black line indicates the current measurement. The blue,
     green, magenta, and red lines correspond to the RDL with short-term memory,
     LSTM, model-based approach, and RDL-MD scheme, respectively. Triangles and
     circles on the time axis mark the start and end points of joint rotations.}
    \label{fig:MD_error_compare}
    \vspace{-3pt}
\end{figure}

The test set included 40\% hysteresis-free (continuous) and 60\% hysteresis-rich
(interrupted) trajectories, totaling 94,490 frames. Qualitative results are
visualized in Fig. \ref{fig:MD_error_compare}, while Table II(a) provides a
quantitative error comparison. Notice that the unit for motor current -
percentage use (\%Use) of loading capacity - may be less intuitive. Since the
manufacturer did not provide motor constants, we used external force-sensing
devices to roughly estimate equivalent torques, as detailed in Section VI. For
the sake of accuracy comparison, we have adhered to using \%Use in this section.

In Fig. \ref{fig:MD_error_compare}, triangle and circle markers on the time axis
signify when a joint starts and stops rotating, respectively. Lacking temporal
information, the RDL model failed to capture hysteresis, leading to significant
errors in static states (e.g., after 6s and 26s, for almost all joints). This
limitation was also evident in the model-based identification method, despite
its effectiveness in dynamic situations. The issue was particularly obvious for
Joints 1, 4, and 6, where both methods inaccurately predicted near-zero values
due to gravity's absence. Conversely, the LSTM model, equipped with non-local
memory, somewhat captured hysteresis but occasionally produced unreliable
drifting estimates in static conditions. Finally, integration of the MD input
scheme with the RDL model ensured consistent and accurate hysteresis
approximation across all scenarios.

\begin{table}[t]
    \begin{subtable}{1.0\linewidth}\centering
        \setlength{\tabcolsep}{5.5pt}
        \begin{tabular}{c c|c c c c c c}
            \hline
            \multicolumn{8}{c}{RMSE (\%Use) Comparison across Models}
            \\ \hline
            Method & L-T-Info & J1 & J2 & J3 & J4 & J5 & J6 \\ \hline
            Model-based & No & 3.92 & 5.30 & 3.90 & 9.60 & 6.49 & 8.08 \\ 
            Basic MLP & No & 5.33 & 5.29 & 4.86 & 9.58 & 4.90 & 7.16 \\
            RDL & No & 3.87 & 4.31 & 3.78 & 9.37 & 4.72 & 7.13 \\
            LSTM & Yes & 5.72 & 4.88 & 4.57 & 6.78 & 3.64 & 5.76 \\
            RDL-MD & Yes & \bfseries 2.49 & \bfseries 3.90 & 3.49 & \bfseries
            5.77 & \bfseries 3.56 & 3.55 \\
            HRDL-MD & Yes & 2.64 & 4.02 & \bfseries 3.18 & 5.84 & 3.73 &
            \bfseries 3.53 \\ \hline
        \end{tabular}
        \vspace{-4pt}
        \caption{}
        \label{table:2a}
    \end{subtable}
    \vspace{2pt}
    
    \begin{subtable}{1.0\linewidth}\centering
        \begin{tabular}{c|c c c c c}
            \hline 
            \multicolumn{6}{c}{Inference Time (ms)} \\ \hline
            & Basic MLP & RDL & LSTM & RDL-MD & HRDL-MD \\ \hline
            Time & 1.45 & 1.46 & 2.21 & 1.75 & \bfseries 0.77 \\ \hline 
        \end{tabular}
        \vspace{-4pt}
        \caption{}
        \label{table:2b}
    \end{subtable}
    \vspace{2pt}

    \begin{subtable}{1.0\linewidth}\centering
        \begin{tabular}{c|c c c c c c}
            \hline
            \multicolumn{7}{c}{RMSE (\%Use) Comparison across Hierarchies} \\
            \hline
            Hierarchy & J1 & J2 & J3 & J4 & J5 & J6 \\ \hline
            1 & 3.38 & 4.54 & 3.77 & 5.97 & 4.06 & 4.20 \\
            2 & 2.70 & 4.10 & 3.20 & 5.87 & 3.73 & 3.60 \\
            3 & \bfseries 2.64 & \bfseries 4.02 & \bfseries 3.18 & \bfseries
            5.84 & \bfseries 3.73 & \bfseries 3.53 \\ \hline
        \end{tabular}
        \vspace{-4pt}
        \caption{}
        \label{table:2c}
    \end{subtable}
    \vspace{-5pt}

    \caption{Error analysis on a test data comprising 94,490 frames, which
    includes both hysteresis-free and hysteresis-rich trajectories. The
    abbreviation "L-T-Info" stands for Long-term Temporal Information.}
    \label{table:model_RMSE}
    \vspace{-12pt}
\end{table}

As mentioned in Section III, the static hysteresis in one joint can be
influenced by the motion states of other joints. This is exemplified in Fig.
\ref{fig:MD_error_compare}, where Joint 4 is in motion from 31s to 34s. Non-zero
hysteresis measurement was present prior to 31s but subsided after 34s. This
behavior could be attributed to the ongoing movements of Joints 2 and 3 beyond
34s, which seemingly counteract the hysteresis in Joint 4 through non-trivial
mechanisms. Similar cases can also be found for Joint 1 at 22s, Joint 5 at 12s,
and Joint 6 at 54s. In such contexts, conventional model-based methods
\cite{liu2021sensorless2}, which consider each joint in isolation, is suboptimal
as it still yield undesirable non-zero hysteresis approximation.

From Table II(a), it's evident that the inclusion of long-term temporal
information significantly reduces RMSE, particularly for Joints 1, 4, 5, and 6.
A comparison between Table II(a) and II(b) reveals that the HRDL-MD scheme
achieves the accuracy of the basic RDL-MD model but with quicker inference,
thanks to the parallel forward propagation design. Further, Table II(c)
indicates that the first hierarchy captures a significant portion of the robot
dynamics, with each subsequent hierarchy further reducing error. These findings
suggest that a shallow HRDL may be adequate for tasks like collision detection,
which require only basic sensing capabilities. On the other hand, tasks that
require a higher degree of accuracy, such as wrench estimation for force
control, may benefit from a deeper HRDL model.

\section{Applications}
\subsection{Joint Compliance with Deadzone Compensation}
Joint compliance is vital for facilitating safe and effective human-robot
interactions, especially in collaborative settings where a robot needs to
dynamically respond to forces applied at unpredictable locations on its body. In
the case of position-controlled robots like the Denso-VS060, users are limited
in their ability to directly manipulate joint torque or current. A commonly
adopted solution, described in \cite{de2006collision}, involves using an
admittance controller, wherein command velocities are set proportional to the
residuals as expressed by $\dot{q}_n=K_p(y_n-y^{est}_n)$. Although the MD input
design substantially mitigates static hysteresis, residual errors from noise and
unmodeled dynamics remain. To ensure reliable admittance control, a deadzone is
introduced for effective error compensation. Similar methodologies are discussed
in \cite{gaz2018model}, which suggests a 10Nm threshold for all joints. However,
such a threshold can make the robot feel `heavy' by activating joints only when
large residuals are present. Consequently, it's essential to empirically
determine the deadzone thresholds to meet two key criteria: halting the robot
immediately upon removal of external forces and making the robot feel possibly
`light'.

\begin{table}[t]
    \vspace{5pt}
    \centering
    \begin{tabular}{c|c c c c c c}
        \hline 
        \multicolumn{7}{c}{Deadzone Boundaries} \\ \hline
        & J1 & J2 & J3 & J4 & J5 & J6 \\ \hline
        Current (\%Use) & 6.0 & 6.0 & 6.0 & 9.0 & 11.0 & 11.0 \\ 
        Torque (Nm) & 4.26 & 4.86 & 2.42 & 1.01 & 1.58 & 0.84 \\ \hline 
    \end{tabular}
    \caption{The deadzone boundaries of current and the equivalent torque
    calculated with external devices.}
    \label{table:3}
    \vspace{-15pt}
\end{table}

The determined deadzone thresholds are presented in Table \ref{table:3}. To
allow for smoother motion transitions, we have incorporated a linear transition
band around each deadzone boundary, effectively mitigating jerky trajectories.
We estimated equivalent torque values using end-effector force measurements at
specific static postures, taking into account the robot's geometry. These values
can also be viewed as the maximum error across all examined joint
configurations.

\begin{figure}[t]
    \vspace{5pt}
    \centering
    \includegraphics[width=0.95\linewidth]{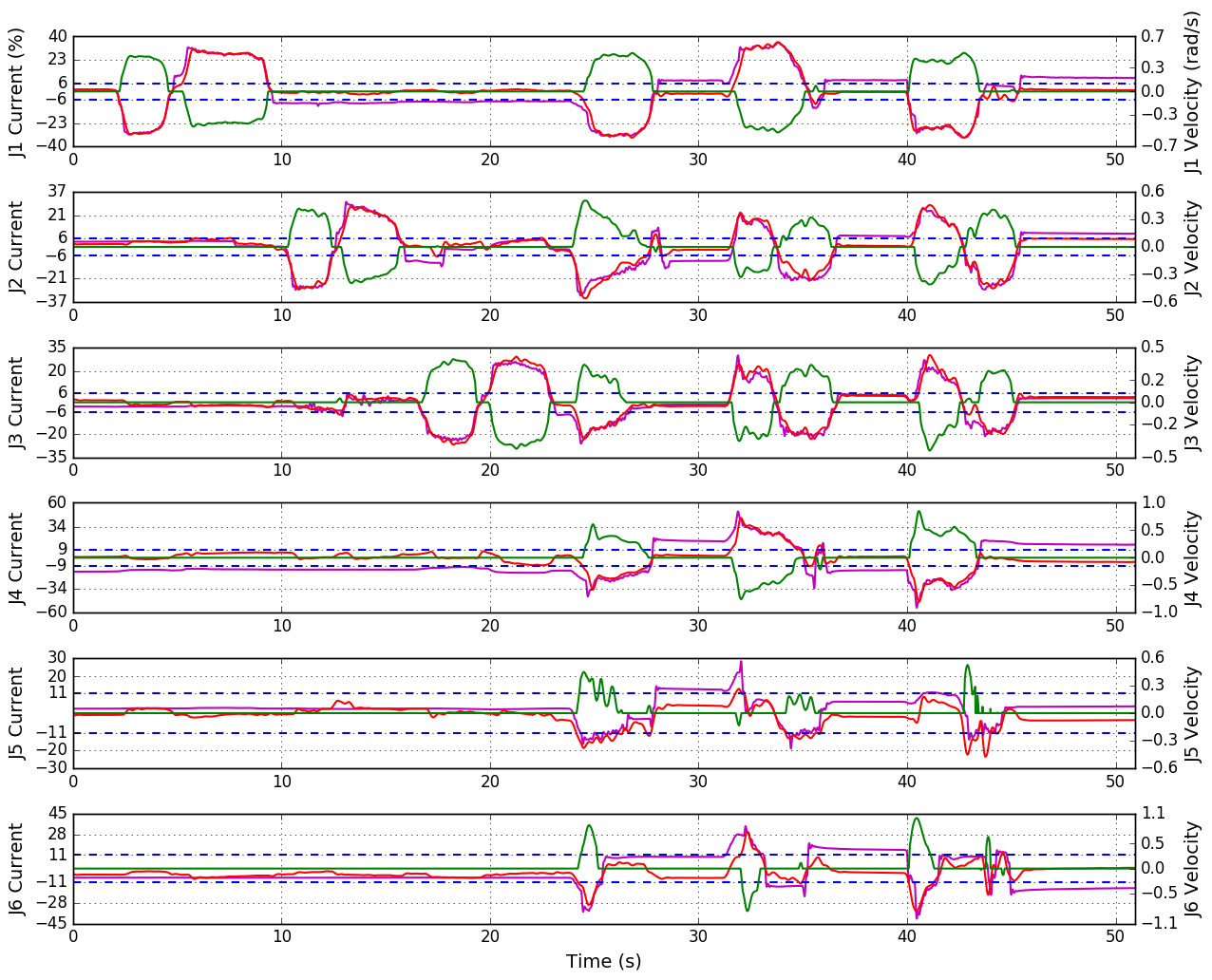}
    \vspace{-5pt}
    \caption{Filtered residuals in joint compliance task. The red and magenta
    lines represent the residuals estimated by the proposed HRDL-MD scheme and
    GMO, respectively. Green lines indicate command velocities, and dashed blue
    lines mark the deadzone boundaries.}
    \label{fig:compliance_compare}
    \vspace{-8pt}
\end{figure}

A joint compliance experiment was subsequently carried out using the HRDL-MD
estimation. We applied appropriate forces to the robot body, moving each link
first separately and then in combination. For safety, constraints were placed on
the command joint positions and velocities. The filtered residuals and
corresponding command velocities are illustrated in Fig.
\ref{fig:compliance_compare}. GMO estimations are also shown for comparative
purposes. It's evident that the GMO scheme without temporal information failed
to identify hysteresis, resulting in non-zero residual predictions even in
contact-free scenarios. To eliminate these inaccuracies, a wider deadzone would
be required, making the robot feel `heavier'.

\subsection{Task Compliance with Wrench Estimation}

\begin{figure}[t]
    \vspace{5pt}
    \centering
    \includegraphics[width=0.95\linewidth]{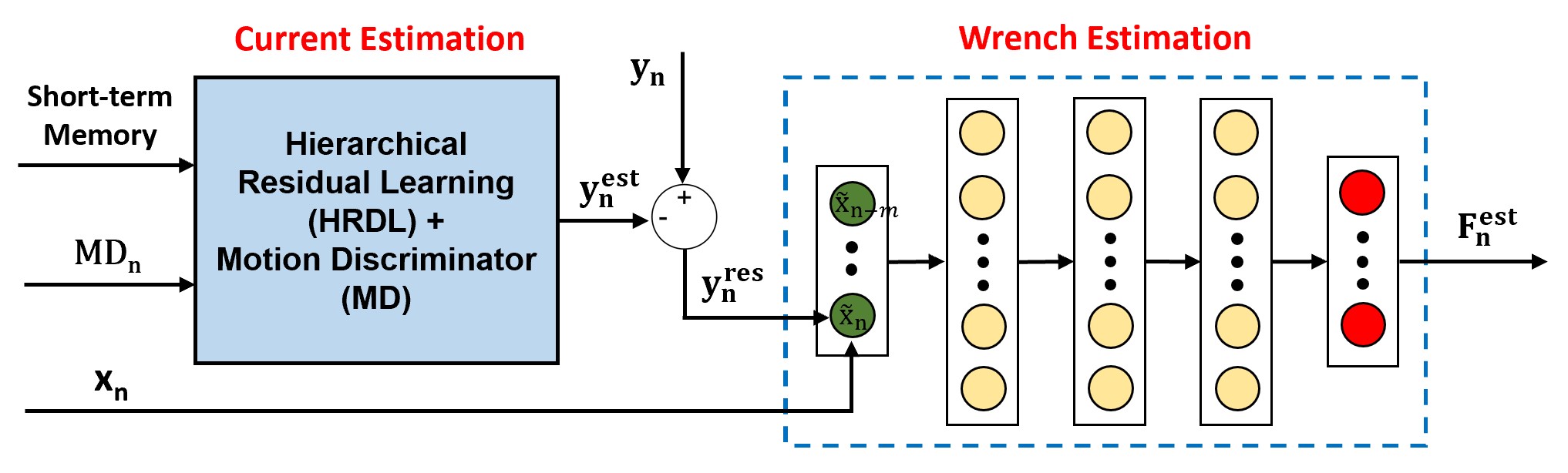}
    \vspace{-5pt}
    \caption{The compound wrench estimation scheme}
    \label{fig:WE_diagram}
    \vspace{-7pt}
\end{figure}

Utilizing the current residuals from our HRDL-MD scheme, one could calculate the
end-effector wrench via the robot's Jacobian. However, given the noise in
current data and the absence of motor constant, we opt for a nonlinear mapping
through an additional MLP model. A related study cited in \cite{10354486}
directly used instantaneous current measurements and joint states as Neural
Network inputs to estimate external wrench. This method yielded accurate and
reliable results in sensorless tight pin insertion tasks. Here, we introduce a
compound Neural Network model, as depicted in Fig. \ref{fig:WE_diagram}, which
uses the current residuals as its input instead of the raw current measurements.

\begin{figure}[t]
    \centering
    \includegraphics[width=0.95\linewidth]{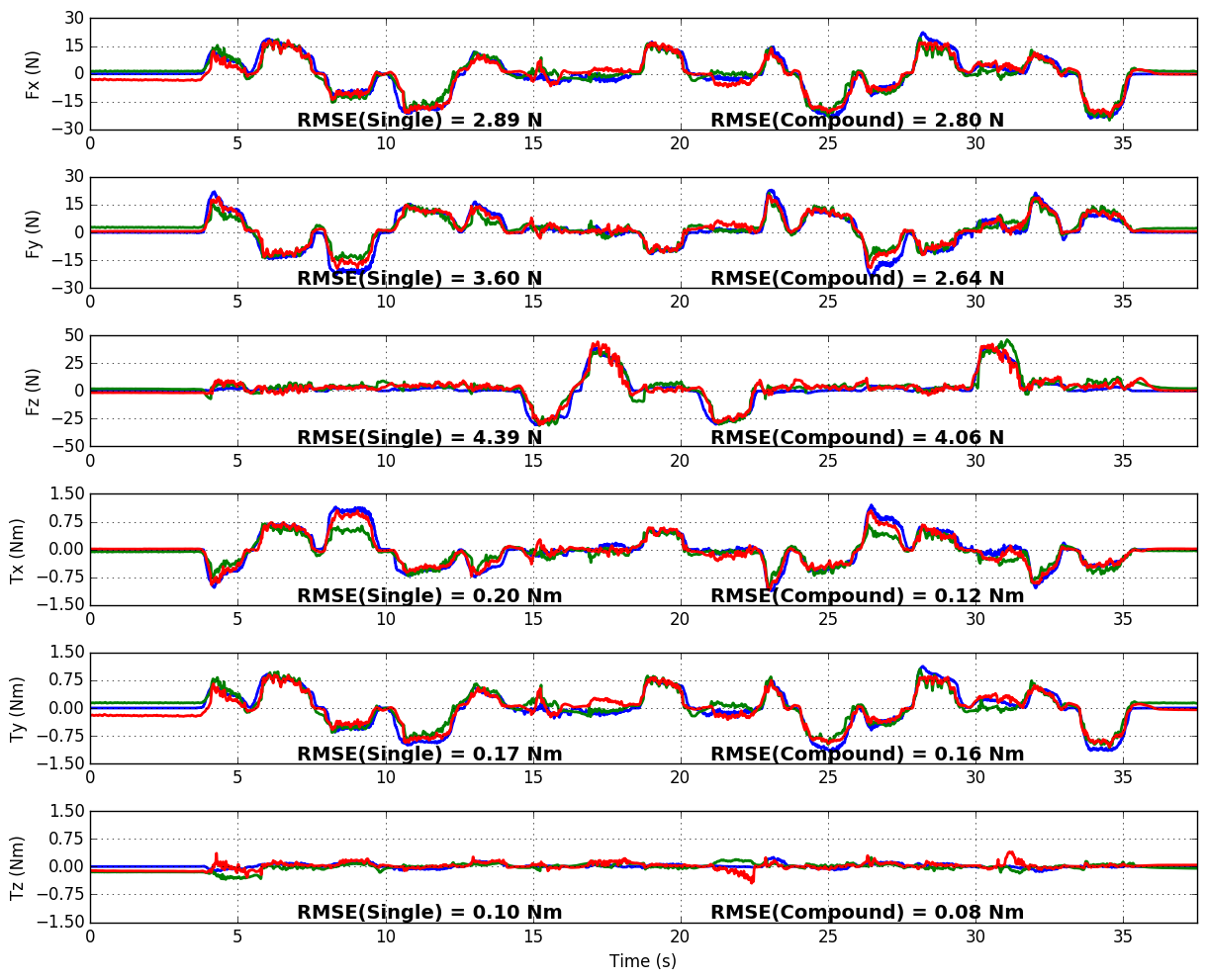}
    \vspace{-5pt}
    \caption{Comparison among the F/T sensor measurement (blue), single model
     estimation (green) and compound model estimation (red) in task compliance.}
    \label{fig:FT_compare}
    \vspace{-11pt}
\end{figure}

We conducted a task compliance experiment with a fixed end-effector orientation,
using an admittance control scheme in the task space and a carefully chosen
deadzone for initiating motion. Figure \ref{fig:FT_compare} presents a
comparison of two methods along with their respective RMSE values. Notably, the
compound-model approach outperforms the single-model method, despite both having
the same number of parameters for wrench estimation. The compound-model is
particularly effective at capturing peak values more accurately. This
improvement can be attributed to the more straightforward residual information
extraction by the HRDL-MD scheme.

Interestingly, the hysteresis phenomenon did not significantly impact task
compliance. In this case, even the single-model method, which did not
incorporate MD input, can make accurate wrench estimations in static states.
This contrasts with other types of robot movements, like decoupled joint
rotations, where the baseline method still displayed hysteresis-like errors in
estimated wrench. However, the compound Neural Network approach was free of such
errors, benefiting from hysteresis compensation at the joint level.

\section{Conclusion}

In this paper, we introduced a Neural Network-based approach for dynamic
identification using motor current measurement. We designed a Motion
Discriminator input scheme, inspired by the Maxwell Slip friction model, to
capture long-term joint state information and effectively address the static
hysteresis problem. Furthermore, we presented a Hierarchical Residual Learning
architecture that enhances estimation accuracy while ensuring short inference
times. Our proposed methods have demonstrated promising results in both joint
and task compliance scenarios, particularly when complemented by well-formulated
control laws and carefully designed deadzones.

One limitation of our study is that the proposed method was implemented under
consistent robot dynamics across data collection, training, and validation
phases. As a result, the Neural Networks model lacks adaptability to dynamic
changes, potentially resulting in inaccurate estimations when a different
end-effector is equipped or external devices are attached to the robot. Future
work will focus on developing a model fine-tuning or calibration scheme to make
the trained model responsive to environmental changes.

\bibliographystyle{ieeetr}
\bibliography{reference} 

\end{document}